\documentclass[journal]{IEEEtran}
\usepackage{cite,url,comment,soul}
\usepackage{xcolor,ulem,multirow,stfloats}
\usepackage[ruled,vlined]{algorithm2e}
\usepackage[pdftex]{graphicx}
\graphicspath{{figures/}}
\usepackage[caption=false,font=footnotesize]{subfig}
\usepackage[cmex10]{amsmath}
\newcommand{\red}[1]{\textcolor{red}{#1}}
\newcommand{\textbu}[1]{\textbf{\underline{#1}}}
\newcommand{\fracpartial}[2]{\frac{\partial #1}{\partial #2}}

\begin{document}
\bstctlcite{IEEEexample:BSTcontrol}
\title{Binarizing Split Learning for Data Privacy Enhancement and Computation Reduction}
\author{Ngoc Duy Pham, Alsharif Abuadbba, Yansong Gao, Tran Khoa Phan, Naveen Chilamkurti
\thanks{Pham, Phan, and Chilamkurti are with School of Computing, Engineering, and Mathematical Sciences (SCEMS), La Trobe University, Victoria, Australia. Email: {\{ngocduy.pham,k.phan,n.chilamkurti\}@latrobe.edu.au}} \thanks{Abuadbba is with CSIRO's Data61 \& Cybersecurity CRC, Australia. Email: sharif.abuadbba@data61.csiro.au}
\thanks{Gao is with Nanjing University of Science and Technology, China. Email: yansong.gao@njust.edu.cn}}
\maketitle
\begin{abstract}
Split learning (SL) enables data privacy preservation by allowing clients to collaboratively train a deep learning model with the server without sharing raw data. However, SL still has limitations such as potential data privacy leakage and high computation at clients. In this study, we propose to binarize the SL local layers for faster computation (up to $17.5$ times less forward-propagation time in both training and inference phases on mobile devices) 
and reduced memory usage (up to $32$ times less memory and bandwidth requirements). More importantly, the binarized SL (B-SL) model can reduce privacy leakage from SL smashed data with merely a small degradation in model accuracy. To further enhance the privacy preservation, we also propose two novel approaches: 1) training with additional local leak loss and 2) applying differential privacy, which could be integrated separately or concurrently into the B-SL model. Experimental results with different datasets have affirmed the advantages of the B-SL models compared with several benchmark models. The effectiveness of B-SL models against feature-space hijacking attack (FSHA) is also illustrated. Our results have demonstrated B-SL models are promising for lightweight IoT/mobile applications with high privacy-preservation requirements such as mobile healthcare applications.
\end{abstract}
\begin{IEEEkeywords}
Split learning, Binarization, Privacy preservation, Leakage loss, Differential privacy.
\end{IEEEkeywords}
\section{Introduction}
\IEEEPARstart{D}{eep} learning (DL) has achieved remarkable performance in many domains such as  computer vision, natural language processing, intrusion detection, and anomaly detection \cite{SurveyDNN17Liu, TKDE21Zheng}. The success of DL stems from the amount of high-quality large-scale datasets, the advancement of high-performance computing, and the maturity of open-source DL frameworks \cite{DPModelPublising19Yu}. For effective standard DL, data from source(s) need to be collected and trained centrally, for example, using cloud infrastructure. However, such DL approach poses data security and/or privacy concerns that violate certain rules such as the general data protection regulation (GDPR) and the health insurance portability and accountability act (HIPAA) \cite{HIPAA-Potamus04}. Data privacy is especially crucial as sensing devices such as smartphones, wearable devices, or even laptops have been becoming personalized with sensitive information. Thus, investigating privacy-preserving DL techniques has been an active research area \cite{NoPeekSurvey18Vepakomma,SurveyPrivateML20Gong,PrivacyDLSurvey20Mireshghallah}. 

For privacy preservation in DL, various approaches have been developed such as differential privacy (DP) mechanisms, homomorphic encryption (HE), secure multi-party computation (SMC), model splitting, partial parameters sharing, mimic learning \cite{PrivacyPreserving20Boulemtafes} with their own advantages and shortcomings. For instance, encryption-based techniques, such as HE and SMC \cite{PrivacyDLSurvey20Mireshghallah}, allow computation over encrypted data which suffers from substantially increased computation overhead and latency, rendering likely unacceptable energy consumption \cite{Scalpel17Yu}. In addition, these techniques have some limitations in terms of the functions they can support \cite{PrivacyDLSurvey20Mireshghallah}. While DP techniques  can provide a provable privacy guarantee for individuals \cite{SurveyPrivateML20Gong}, it requires a balance between privacy and accuracy when applying DP to DL despite research efforts to alleviate the impact of noise perturbation on the accuracy of the inference phase \cite{NotJustPrivacy18Wang}. To avoid the undesired centralized data collection, federated learning (FL) allows local data processing without the need to transfer data to the server. In FL, local models are trained on local data before their parameters are exchanged to aggregate a global model. However, training a whole DL model locally can create a large overhead for low-end resource-constrained devices. Besides, it has been shown that there are potential information leaks by observing parameter gradients of FL to conduct various privacy inference attacks (i.e. membership inference attack, data inversion attack, and preference profiling attack) \cite{DeepLeakage19Zhu,InvertGradients20Geiping,zhou2022ppa}. 

Keeping distributed setting but relaxing full local training, split learning (SL) splits and allocates layers to both the client and the central server. The local client performs the first few layers and the computation of rest majority layers are offloaded to the more powerful server. This is particularly attractive for IoT as the latest generation DL models consist of million up to hundred million parameters that require enormous computing ability to train and process \cite{NotJustPrivacy18Wang}. Compared with FL, in SL, partial model weights stay private, hence white-box model inversion attacks on clients are not possible. Messaging and network overhead of SL are significantly less than FL especially for large  model since model parameters never leave the client. Nevertheless, the computation overhead from industry-standard models or communication during learning between clients and server in SL is still expensive for resource-limited clients \cite{SplitComputing21Matsubara}. Second, it has been shown that SL suffers from high data leakage in the case of time-series data \cite{SL1DCNN20Sharif}. There have been  progresses to address these limitations of SL. 

\begin{figure*}[t]
\centering
\includegraphics[width=0.7\textwidth]{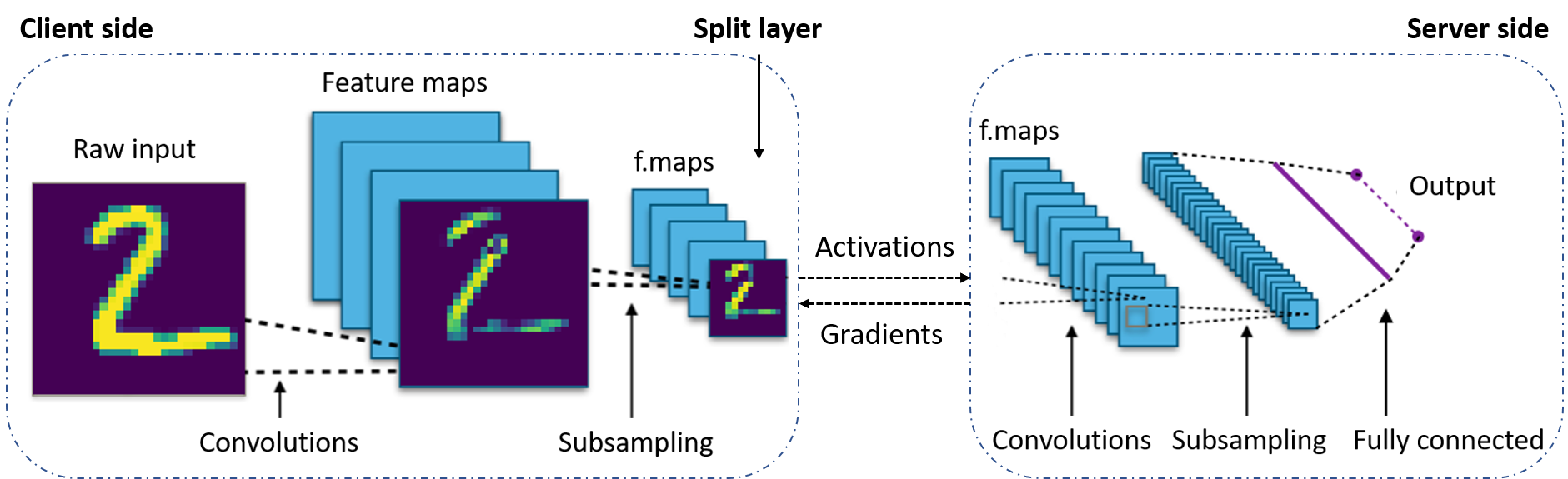}
\caption{Visual invertibility of a typical CNN under SL setting. The feature maps show images visually plotted from a channel in the first convolution and subsampling layers.}
\label{fig:CNN_Leakage}
\end{figure*}

\begin{figure}[h]
    \centering
    \includegraphics[width=0.3\textwidth]{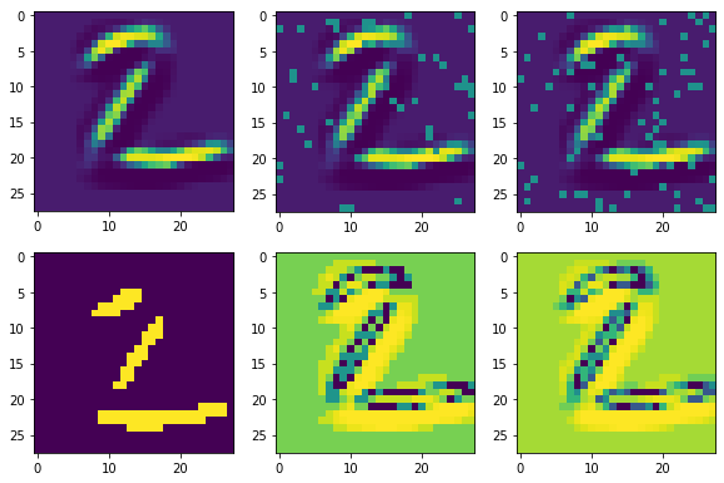}
    \caption{Visual invertibility of \cite{DroppingActivations18Dong} and \cite{StepwiseActivations19Yu} approaches from our experiments. The top row, from left to right, shows the visually plotted images from activations of a convolution layer using Dropping activation outputs with $p$ set to $0$, $5$, and $9$ \cite{DroppingActivations18Dong}. The bottom row shows the reconstruction results using Stepwise activation function with $v=10$, $n$ set to $2$, $11$ and $21$ \cite{StepwiseActivations19Yu}.}
    \label{fig:Related_Work}
\end{figure}

Recently, Abuadbba et al. \cite{SL1DCNN20Sharif} have studied the application of SL to perform privacy-preserving training for ECG health data. The authors have demonstrated that pure SL on 1D-CNN may fail to protect raw data privacy. Fig.  \ref{fig:CNN_Leakage} exhibits the potential leakage of raw data at the split layer under a typical CNN model. With a similar slitting manner, Dong et al. \cite{DroppingActivations18Dong,DroppingActivations20Xinrui} have proposed to split the deep network into two parts and localize only the first hidden layer. The authors then apply the \textit{Dropping activation outputs} method to the local layer's activations to make them non-invertible to prevent leakage. However, when applying their method to the first convolution layer of a CNN, leakage could happen when visually plotting feature maps as we experimentally demonstrated in Fig. \ref{fig:Related_Work}. Likewise, Yu et al. \cite{StepwiseActivations19Yu} also partition the model after the first layer and apply the proposed \textit{Stepwise activation function} method to make activation outputs irreversible. The effectiveness of their proposal is proportional to the Stepwise parameters and it also exhibits a trade-off between accuracy and privacy preservation. Studying another approach, in NoPeek \cite{NoPeek20Vepakomma,ReducingLeakage19Vepakomma}, Vepakomma et al. have shown that by reducing the distance correlation between the intermediary representations and raw data, information leakage from reconstruction attacks could be prevented. The authors focus on distance correlation metric and white-box reconstruction attacks without considering direct leakage from visual invertibility. In additions, NoPeek is ineffective against the feature-space hijacking attack (FSHA) \cite{FSHA_SL21Pasquini} which is a state-of-the-art black-box attack aiming to reconstruct local private raw data. 

\textit{Our contributions.} Taking into account the privacy preservation and computation issues in SL, our study focuses on the development of novel SL models to reduce data leaks with low computation overhead by leveraging Binarized Neural Network (BNN) \cite{BinarizedNN16Courbariaux,BinarizedNN16Hubara}. BNN is hardware-friendly in terms of computation costs and memory usage with binary weights and activations. Binarization, an extreme quantization, causes loss of information which consequently is a potential technique to ensure privacy \cite{NoiselessPrivacy19Farokhi}.  In summary, the main contributions to this paper are:

\begin{itemize}
\item We propose B-SL to binarize only the local part of SL to make the model light on the clients, but can still achieve accuracy comparable to those of the full precision models. We rigorously analyze the computation overhead and privacy preservation capability of the proposed approach. 
\item We propose a leakage-restriction training strategy to further reduce data leaks. This method combines leak loss of local private data and main accuracy loss that enhance privacy while maintaining model utility.
\item We propose three approaches to integrate DP into B-SL and prove that binarization naturally satisfies $(\epsilon,\delta)$-DP \cite{ApproximateDP06Dwork} with a bounded $\epsilon$ corresponding to a given $\delta$. 
\item We present a comprehensive empirical evaluation of the proposed B-SL models on standard image classification tasks and compare the achieved results with those of four other baselines. We also experimentally demonstrate the effectiveness of  B-SL against FHSA. Our code is open-sourced to GitHub\footnote{https://github.com/phamngocduy/BinarizeLocalizedLayers}.
\end{itemize}

The remainder of this paper is organized as follows: Section \ref{sec:background} provides background information about SL and BNN. The design, implementation, and analysis of B-SL are detailed in Section \ref{sec:binarized}. Two proposed techniques to further enhance privacy preservation for B-SL are described in Section \ref{sec:privacy}. Section \ref{sec:experiment} describes the experimental results in terms of accuracy and privacy of the proposed B-SL and other benchmark models followed by the conclusion in Section \ref{sec:conclusion}.
\section{Background} \label{sec:background}
This section provides the background information for SL and BNN, which is the stepping stone for our proposed B-SL.

\subsection{Split Learning}
\begin{figure}[h]
    \centering
    \includegraphics[width=0.45\textwidth]{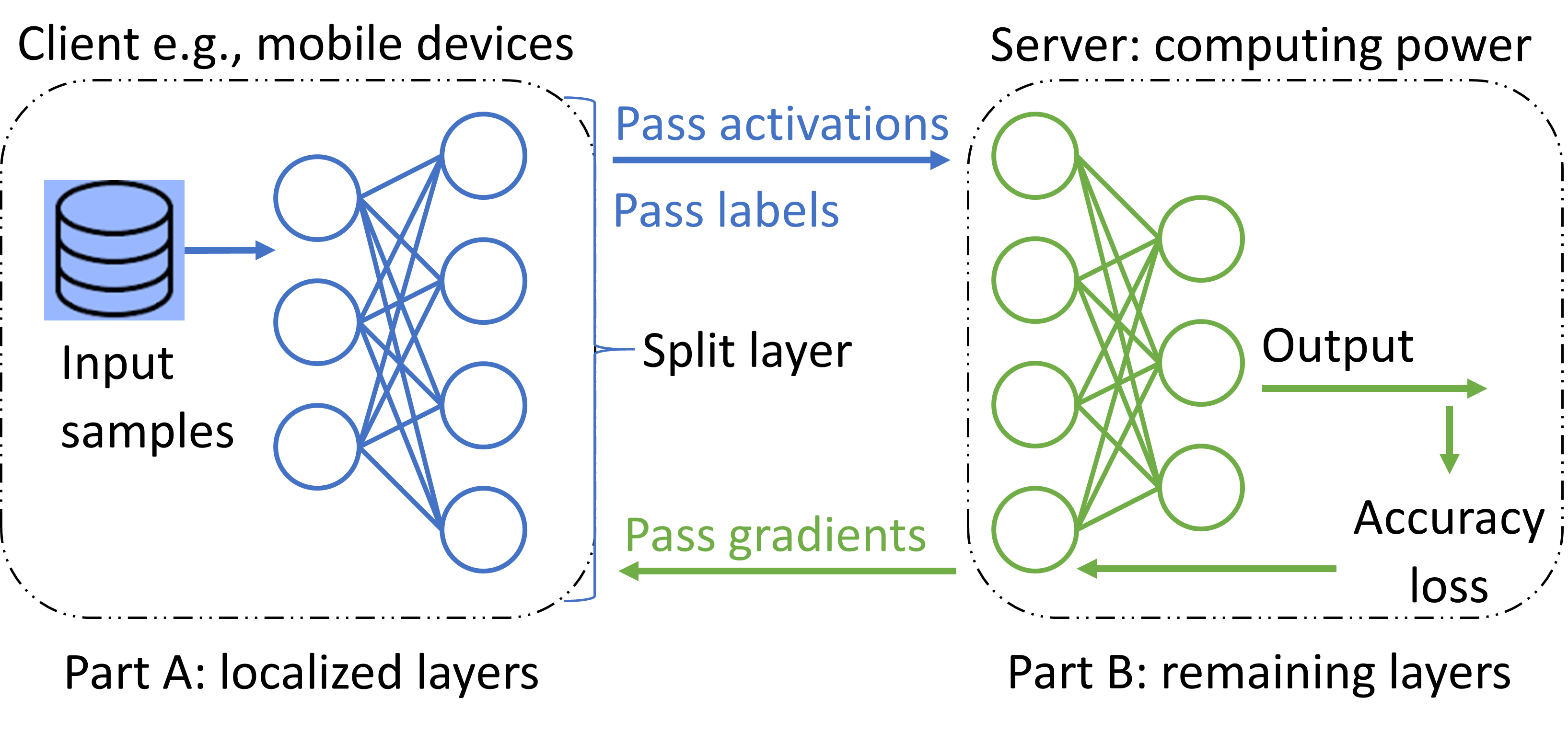}
    \caption{SL network architecture with two parties.}
    \label{fig:SLArchitecture}
\end{figure}

Split learning  splits a DL model into multiple parts which are trained on different entities \cite{DistributedLearning18Gupta}. The training of the network is carried forward by transferring the activations of the last layer of each part (also known as the split layer) to the next part. Thus, no raw data is shared among entities except split layer's activations. SL is considered and validated for multi-dimensional (e.g. 1D, 2D) models with various architectures such as LeNet, VGG, AlexNet \cite{SL1DCNN20Sharif,DistributedLearning18Gupta}. Fig. \ref{fig:SLArchitecture} depicts a simplified vanilla configuration of SL between a client (e.g., a smart camera continuously collects environmental information) and a  server.

\begin{algorithm}[t]
\SetAlgoLined
$s\leftarrow$ client socket initialized\\
$s.connectAndSyncParams(Server)$\\
\For{each batch $(x,y)$} {
    \textbf{Forward propagation:}\\
    $a^{(0)}\leftarrow x$\\
    \For{$i\leftarrow 1$ to $l$}{
        $z^{(i)}\leftarrow f^{(i)}(a^{(i-1)},w^{(i)})$\\
        $a^{(i)}\leftarrow g^{(i)}(z^{(i)})$\\
    }
    $s.send\left((a^{(l)},y)\right)$\\
    \textbf{Backward propagation:}\\
    $\fracpartial{E}{a^{(l)}} \leftarrow s.receive()$\\
    \For{$i\leftarrow l$ downto $1$}{
        $\fracpartial{E}{z^{(i)}}\leftarrow\fracpartial{E}{a^{(i)}}\times g^{(i)}{'}\left(z^{(i)}\right)$\\
        $\fracpartial{E}{w^{(i)}}\leftarrow\fracpartial{E}{z^{(i)}}\times\fracpartial{z^{(i)}}{w^{(i)}}$\\
        \uIf{$i \neq 1$}{
            $\fracpartial{E}{a^{(i-1)}}\leftarrow\fracpartial{E}{z^{(i)}}\times\fracpartial{z^{(i)}}{a^{(i-1)}}$\\
        }
        Update $w^{(i)}$ using $\fracpartial{E}{w^{(i)}}$
    }
}
$s.close()$
\caption{SL process on the client.}
\label{alg:client}
\end{algorithm}

Consider a deep model that has $L$ layers excluding the input layer and assume that we split the model after layer $(l)$. The first $l$ layers from $(1)$ to $(l)$ are deployed in the client and the remaining layers from $(l+1)$ to $(L)$ are in the server. Denote $w^{(i)}$, $f^{(i)}$, $z^{(i)}$, and $a^{(i)}$ as weights, forward function, output tensors, and activation output at layer $(i)$, respectively. The procedures for processing learning on each side are presented in Algorithms \ref{alg:client} and \ref{alg:server}. The SL process starts on the client following Alg. \ref{alg:client}.

\begin{enumerate}
\item \textbf{Client}. Firstly, the client connects to the server for synchronizing training hyper-parameters (e.g. batch size, learning rate, \dots). For each batch, the client processes forward propagation up to layer $(l)$ and then sends $a^{(l)}$ to the server. Later, on receiving $\fracpartial{E}{a^{(l)}}$ (where $E$ is the model's loss) from the server, the gradients are back-propagated from $(l)$ to the first hidden layer.

\item \textbf{Server}. As indicated in Alg. \ref{alg:server}, the server conducts a traditional learning process by continuing the forward propagation after receiving activations $a^{(l)}$ from the client. Subsequently, the server calculates the loss $E$ between the model outputs and labels passed by the client. Then, gradients calculated from loss $E$ are back-propagated from the last layer $(L)$ until $(l+1)$ followed by sending $\fracpartial{E}{a^{(l)}}$ to the client for continuing the back-propagation locally.
\end{enumerate}
\begin{algorithm}[h]
\SetAlgoLined
$s\leftarrow$ server socket initialized\\
$s_C\leftarrow s.acceptAndSyncParams(Client)$\\
\For{$i\leftarrow 1$ to $N$} {
    \textbf{Forward propagation:}\\
    $(a^{(l)},y)\leftarrow s_C.receive()$\\
    \For{$i\leftarrow l+1$ to $L$}{
        $z^{(i)}\leftarrow f^{(i)}(a^{(i-1)},w^{(i)})$\\
        $a^{(i)}\leftarrow g^{(i)}(z^{(i)})$\\
    }
    $E\leftarrow\mathcal{L}(a^{(L)},y)$\\
    \textbf{Backward propagation:}\\
    Compute $\fracpartial{E}{a^{(L)}}$\\
    \For{$i\leftarrow L$ downto $l+1$}{
        $\fracpartial{E}{z^{(i)}}\leftarrow\fracpartial{E}{a^{(i)}}\times g^{(i)}{'}(z^{(i)})$\\
        $\fracpartial{E}{w^{(i)}}\leftarrow\fracpartial{E}{z^{(i)}}\times\fracpartial{z^{(i)}}{w^{(i)}}$\\
        $\fracpartial{E}{a^{(i-1)}}\leftarrow\fracpartial{E}{z^{(i)}}\times\fracpartial{z^{(i)}}{a^{(i-1)}}$\\
        Update $w^{(i)}$ using $\fracpartial{E}{w^{(i)}}$\\
    }
    $s_C.send\left(\fracpartial{E}{a^{(l)}}\right)$\\
}
$s_C.close()$
\caption{SL process on the server.}
\label{alg:server}
\end{algorithm}
The split layer's activations, which are transmitted from client to server during forwarding propagation, are also called smashed data \cite{SLComparedFL19Abhishek}. As the output of a function (smasher) on raw data, this smashed data results in potential leakage that could be reduced by the proposed binarization of SL in the next section.
\subsection{Binarized Neural Network}
BNNs \cite{BinaryConnect15Courbariaux,BinarizedNN16Courbariaux,BinarizedNN16Hubara} are neural networks where weights and activations are constrained to either $-1$ or $+1$ for mathematical convenience. We can implement BNN using $\{0,1\}$ encoding with XNOR and a few integer operations (i.e. bit-count) which are the most energy-efficient \cite{StochasticBNN@Livochka}. The main drawback of BNNs is that they might not be able to achieve as good accuracy as their full-precision counterparts. However, there have been recent advancements to narrow the accuracy gap \cite{SurveyBNNQin}. Currently, BNN is a good candidate for DL implementations on FPGAs and ASICs \cite{ReviewBNN@Simons}.
\section{Binarized split learning} \label{sec:binarized}
In this section, we propose an algorithm to binarize SL for enhancing privacy preservation and reducing computation overhead and memory usage.
\subsection{Binarized SL (B-SL)}
Consider the SL architecture depicted in Fig. \ref{fig:SLArchitecture}, we propose to binarize the local part of the deep model to speed up computation on the clients. Meanwhile, most of the resource-hungry operations which run on the powerful server are kept in high-precision computation to retain the accuracy of the whole DL model. The proposed B-SL is described in Alg. \ref{alg:client_bin} with all modifications compared to pure SL being highlighted with red color. We use the deterministic \textsf{Sign} function to binarize/transform weights and activations from real values to binary values:

\begin{equation}
    x_b = \textsf{Sign}(x) =
    \begin{cases}
        +1 & \text{if $x \geq 0$}\\
        -1 & \text{otherwise,}
    \end{cases}
\end{equation}

\noindent where $x_b$ is the binarized value of the real-value variable $x$. Because the gradient of the \textsf{Sign} function is zero for almost the values, a solution is using \textit{Straight-Through Estimator} \cite{StraightThroughEstimator13,UnderstandingSTE19Yin} which passes the gradient exactly as it is during back-propagation. This simplifies the back-propagation for threshold (\textsf{Sign}) function in BNN and has shown to work very well. In back-propagation, because real-valued gradients of weights are accumulated in real-valued variables, we need to use real-value weights for Stochastic Gradient Descent to work at all. \textit{Batch Normalization} \cite{BatchNormalization15} and \textit{Weights Clipping} \cite{BinarizedNN16Hubara} are also the key techniques to accelerate the training and to reduce the impact of the weights' scale because real-valued weights would grow very large without any impact on the binary weights. More details on BNNs can be referred to \cite{BinarizedNN16Hubara}. In Alg. \ref{alg:client_bin} where we present the B-SL process on the client, \textsf{BatchNorm} indicates batch normalization and \textsf{BackBatchNorm} specifies how to back-propagate through the \textsf{BatchNorm}. The process of B-SL is as follows:

\begin{algorithm}[t]
\SetAlgoLined
$s\leftarrow$ client socket initialized\\
$s.connectAndSyncParams(Server)$\\
\For{each batch $(x,y)$} {
    \textbf{Forward propagation:}\\
    $a_{\red{b}}^{(0)}\leftarrow x$\\
    \For{$i\leftarrow 1$ to $l$}{
        \red{$w_b^{(i)}\leftarrow\textsf{Sign}\left(w^{(i)}\right)$}\\
        $z^{(i)}\leftarrow f^{(i)}\left(a_{\red{b}}^{(i-1)},w_{\red{b}}^{(i)}\right)$\\
        $a^{(i)}\leftarrow \red{\textsf{BatchNorm}}\left(z^{(i)}\right)$\\
        $\red{a_b^{(i)}\leftarrow\textsf{Sign}\left(a^{(i)}\right)}$
    }
    $s.send\left((a_{\red{b}}^{(l)},y)\right)$\\
    \textbf{Backward propagation:}\\
    $\fracpartial{E}{a_{\red{b}}^{(l)}}\leftarrow s.receive()$\\
    \For{$i\leftarrow l$ downto $1$}{
        \red{$\fracpartial{E}{a^{(i)}}\leftarrow\fracpartial{E}{a_b^{(i)}}\circ{1}_{|a^{(i)}|\leq{1}}$}\\
        $\fracpartial{E}{z^{(i)}}\leftarrow\fracpartial{E}{a^{(i)}}\times\red{\textsf{BackBatchNorm}\left(z^{(i)}\right)}$\\
        $\fracpartial{E}{w_{\red{b}}^{(i)}}\leftarrow\fracpartial{E}{z^{(i)}}\times\fracpartial{z^{(i)}}{w_{\red{b}}^{(i)}}$\\
        \uIf{$i \neq 1$}{
            $\fracpartial{E}{a_{\red{b}}^{(i-1)}}\leftarrow\fracpartial{E}{z^{(i)}}\times\fracpartial{z^{(i)}}{a_{\red{b}}^{(i-1)}}$\\
        }
        Update $w^{(i)}$ using $\fracpartial{E}{w_{\red{b}}^{(i)}}$\\
        $\red{w^{(i)}\leftarrow\textsf{Clip}\left(w^{(i)},-1,1\right)}$
    }
}
$s.close()$
\caption{B-SL process on the client.}
\label{alg:client_bin}
\end{algorithm}

\begin{itemize}
\item \textbf{Client}. After connecting to the server and synchronizing configuration, for each batch, the client processes forward propagation until the split layer $l$. At each layer $i$, we first binarize $w^{(i)}$ to $w_b^{(i)}$ which is used to compute layer's output $z^{(i)}$. Next, we apply a batch normalization followed by a binarization to obtain layer' activations $a_b^{(i)}$ which is fed as input to the next layer until layer $l$. Then, the client sends activations $a_b^{(l)}$ to the server.\\
On receiving $\fracpartial{E}{a_b^{(l)}}$ from the server, the client processes the back-propagation back to the first hidden layer. At each layer $i$, we obtain $\fracpartial{E}{a^{(i)}}$ from $\fracpartial{E}{a_b^{(i)}}$ via straight-through-estimator: $g_r=g_b\circ1_{|r|\leq1}$ \cite{BinarizedNN16Hubara}. $\fracpartial{E}{a^{(i)}}$ is used to compute gradients of layer's weights and previous layer's gradients also. After updating layer's weights, we constrain them into $[-1,1]$ by using the \textsf{Clip} function.
\item \textbf{Server}. The process in the server is kept intact as in learning with the high-precision model (see Alg. \ref{alg:server}). Note that, the server receives binary activations from the client instead of real values as in traditional SL. Despite the server receives the binary values, it will treat them as full-precision values for the server side forward computation.
\end{itemize}
\subsection{Computation and Memory Reduction Analysis of B-SL}
Binarization not only saves the expensive model storage but also reduces the matrix computation costs by using XNOR and bit-count operations. Hubara et al. \cite{BinarizedNN16Hubara} claimed their implemented binary matrix multiplication is $7$ times faster on the GPU than the non-binary version, while Rastegari et al. \cite{XNORNet16Rastegari}  reported that BNN can have $32$ times lower memory saving and $58$ times faster convolution operations than 32-bit CNN. By binarizing the local model of SL, clients would inherit all benefits brought by BNN which is a huge advantage for IoT. 

\begin{figure}[h]
    \centering
    \subfloat[Replicating kernels]{
        \includegraphics[width=0.25\textwidth]{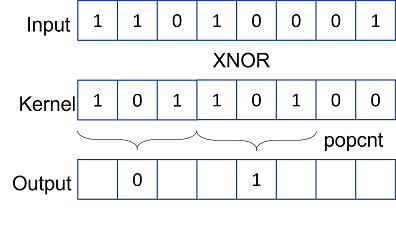}
        \label{fig:Convolution_Speedup}
    }
    \hfill
    \subfloat[Speedup ratio]{
        \includegraphics[width=0.45\textwidth]{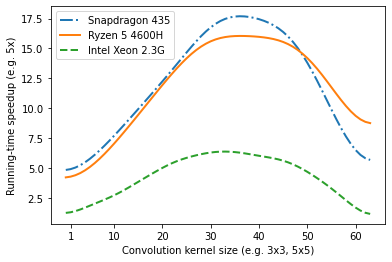}
        \label{fig:RunningTime_Comparison}
    }    
    \caption{Convolution speedup with kernel replication.}
\end{figure}

We affirm the advantages of binarization by comparing the running time during inference of binarization and non-binarization versions of a convolution layer on mobile devices. Besides using bit packing, XNOR operation, and bit-count instruction as in \cite{BinarizedNN16Hubara} and \cite{XNORNet16Rastegari}, with convolution layer we replicate kernel along with a variable to perform multi convolve within an operation as described in Fig. \ref{fig:Convolution_Speedup}. Running on 64-bit architecture devices, Fig. \ref{fig:RunningTime_Comparison} presents the speedup ratio over different kernel sizes. We can observe that on a powerful server (Linux based Intel Xeon CPU 2.3GHz) with integrated floating-point units, the speedup of binarization is up to $7$ times while the binarized convolutions on laptops (Windows based AMD Ryzen 5 4600H) and mobile phones (Android based Qualcomm Snapdragon 435) achieve a higher speedup that is up to $17.5$ times compared to non-binarization. 

With regards to memory storage, by binarization, we replace a 32-bit floating-point number with a 1-bit binary value so the memory saved is clearly 32 times. In addition, if bit-packing technique is used (pack $32$ bits into a $32$-bit integer number), it could save more memory storage. Besides computation overhead, the cost of smashed data transmitting between client and server in SL is also an important concern \cite{SplitComputing21Matsubara}. Therefore, data size saving via binarization ($32$ or $64$ times, depending on processor architecture) is a significant advantage that leads to saving bandwidth/energy and improving the throughput of client-server wireless communication.

It should be noted that during the training phase, backward propagation is still working with real-value gradients. In order to accumulate these gradients, we have to keep the full-precision variables of local layer weights ($w^{(i)}$) which then be binarized ($w_b^{(i)}$) to process forward pass. This extra storage overhead only appears in the training phase; while during the inference phase, only binary weights are needed to achieve a smaller model size with fast forward propagation. Of course, during training phase, we still achieve similar benefit of fast forward propagation and small size activations transferring between client and server.
\subsection{Privacy Preservation Analysis of B-SL}
Although SL has maintained local data untouched from outside, leakage still happens in previous works. In order to analyze the privacy leakage of the B-SL, we define a threat model as follows.

\paragraph{Threat model.} We assume that the server is an honest-but-curious \cite{HonestButCurious14Paverd} adversary, i.e., the server will perform all its operations as specified but is curious about the raw data located on the clients. Such adversarial server tries to reconstruct raw data from smashed data that are delivered by the client during forwarding propagation. 

In the B-SL model, privacy leakage happens when the server can infer raw data from the received smashed data \cite{SL1DCNN20Sharif,NotJustPrivacy18Wang,InversionAttacks19He}. As a popular mitigation technique, Differential Privacy (DP) can provide a formal privacy guarantee for smashed data \cite{SL1DCNN20Sharif,NotJustPrivacy18Wang} in terms of privacy budget $\epsilon$. A common approach to achieve $\epsilon$-DP (a mathematical definition for the privacy loss from \cite{DifferentialPrivacy06Dwork}) for a mechanism $\mathcal{M}$ is adding Laplace noises to the output of a function $f(.)$ such as:
\begin{equation}
\mathcal{M}(x)=f(x)+\textsf{Lap}(s/\epsilon)
\label{eq:lap_noise}
\end{equation}
\noindent where $s$ is the sensitivity of $f$ and \textsf{Lap}$(S)$ denotes sampling from Laplace distribution with center $0$ and scale $S$. As a function on raw input data $x$, if we add noise \textsf{Lap}$(s/\epsilon)$ to the split layer's activations $f(x)$ then the smashed data $\mathcal{M}(x)$ provides $\epsilon$-DP. Based on the Laplace noise adding, we analyze and prove that binarization provides better privacy protection compared to non-binarization as follows.

Considering our proposed B-SL where the smashed data is binarized, we rewrite Eq. \ref{eq:lap_noise} as:
\begin{equation}
\begin{split}
\mathcal{M}_b(x)&=\textsf{Sign}(\mathcal{M}(x))\\
&=\textsf{Sign}\left(\textsf{Sign}(f(x))+\textsf{Lap}(2/\epsilon)\right)
\end{split}
\label{eq:bin_noise}
\end{equation}
\noindent where the sensitivity of the \textsf{Sign} function is $2$. If $|\textsf{Lap}(2/\epsilon)|<1$ then $\mathcal{M}_b(x)=\textsf{Sign}(f(x))$. \textit{It means that binarization naturally contains a small latent noise. Having added noise protects privacy better.} In addition, regarding Eq. \ref{eq:bin_noise}, binarization would provide $\epsilon$-DP with some specific $\epsilon$ values. In order to find the lower bound of these $\epsilon$s, we denote \textsf{cdf} as the cumulative distribution function of Laplace, then the probability of $|\textsf{Lap}(2/\epsilon)|<1$ is calculated as:
\begin{equation}
\begin{split}
    \textsf{Pr}\left[\left|\textsf{Lap}(2/\epsilon)\right|<1\right]&=1-\left[\textsf{cdf}(1)-\textsf{cdf}(-1)\right]\\
    &=1-\exp(-\epsilon/2).
\end{split}
\end{equation}

Regarding the definition of approximate DP, $(\epsilon,\delta)$-DP \cite{ApproximateDP06Dwork}, deterministic binarization could provide $\epsilon$-DP with probability $(1-\exp(-\epsilon/2))$. That means binarization naturally provides $(\epsilon,\delta)$-DP with $\delta=\exp(-\epsilon/2)$; the lower bound of $\epsilon$ is $-2\log(\delta)$. Therefore, \textit{we can conclude that binarization preserves better privacy compared to non-binarization.} In the next section, we investigate further privacy enhancement via investigate binarization with fully $\epsilon$-DP (with any $\epsilon$) together with a novel leakage restriction training.
\section{Privacy-enhancing binarized SL} \label{sec:privacy}
To further reduce privacy leakage, we investigate a number of strategies that could be incorporated into the B-SL model. Specifically, we propose i) leakage restriction training and ii) applying DP on smashed data, which are complementary.
\subsection{Leakage Restriction Training}
Inspired by the device level sanitization scheme \cite{NoPeek20Vepakomma} where the authors proposed to quick decorrelate the distance correlation between the client's raw data and intermediate activations before allowing any communication for training; we propose to exploit an additional loss term along with the model accuracy loss. The additional loss, called leak loss, is used to measure the leakage of local private data. Leak loss is minimized during training to further reduce the leakage from smashed data before being delivered to the server. The leak loss in our approach is general that could be any leakage metric instead of the specific distance correlation in \cite{NoPeek20Vepakomma}.

\begin{algorithm}[t]
\SetAlgoLined
$s\leftarrow$ client socket initialized\\
$s.connectAndSyncParams(Server)$\\
\For{each batch $(x,y)$} {
    \textbf{Forward propagation:}\\
    $a_b^{(0)} \leftarrow x$\\
    \For{$i \leftarrow 1$ to $l$}{
        \textit{Same as in {\bf{Algorithm \ref{alg:client_bin}}}}
    }
    $s.send\left((a_b^{(l)},y)\right)$\\
    $\red{E_2\leftarrow\mathcal{L}_2\left(a_b^{(l)},x\right)}$\\
    \textbf{Backward propagation:}\\
    $\fracpartial{E_{\red{1}}}{a_b^{(l)}}\leftarrow s.receive()$\\
    $\red{\fracpartial{E}{a_b^{(l)}}\leftarrow\lambda\fracpartial{E_1}{a_b^{(l)}}+(1-\lambda)\fracpartial{E_2}{a_b^{(l)}}}$\\
    \For{$i\leftarrow l$ downto $1$}{
        \textit{Same as in {\bf{Algorithm \ref{alg:client_bin}}}}\\
        \red{Update $w^{(i)}$ using Eq. (\ref{eq:update_w})}
    }
}
$s.close()$
\caption{B-SL with leakage restriction (procedure on the client).}
\label{alg:client_leak}
\end{algorithm}

The total loss function for input data $x$, activations $a^{(l)}$ of split layer, predicted labels $a^{(L)}$, true labels $y$ and a scalar weights $\lambda$ is  given by:
\begin{equation}
\begin{split}
    \mathcal{L}_{total} &=\lambda E_1 + (1-\lambda)E_2 \\
    &=\lambda \ \underset{W_1,W_2}{\mathcal{L}_1}(a^{(L)},y) + (1-\lambda) \ \underset{W_1}{\mathcal{L}_2}(a^{(l)},x)
\end{split}
\label{eq:Total_Loss}
\end{equation}

\noindent where $\mathcal{L}_1$ is the main accuracy loss of the deep model. $\mathcal{L}_2$ is the leak loss which can be any leakage measure that we want to optimize. The model parameters $W$ ($w^{(1)} \dots w^{(L)}$) is split as $W_1$ ($w^{(1)} \dots w^{(l)}$) and $W_2$ ($w^{(l+1)} \dots w^{(L)}$) which are distributed to partial models on the client and the server, respectively. From  Eq. (\ref{eq:Total_Loss}), $\mathcal{L}_2$ is calculated on the basis of local data and is influenced only by these data. Therefore, we localize the calculating leak loss and the step of updating gradient based on this loss to the device. After sending activations to the server, the client calculates the loss $E_2=\mathcal{L}_2(a^{(l)},x)$. Then, upon computing $\frac{\partial E_1}{\partial w^{(l)}}$ from the received $\frac{\partial E_1}{\partial a^{(l)}}$, along with the locally computed gradient $\frac{\partial E_2}{\partial w^{(l)}}$, the client updates its local parameters as following:

\begin{equation}
    w^{(i)} = w^{(i)} - \eta\left(\lambda\frac{\partial E_1}{\partial w^{(i)}} + (1-\lambda)\frac{\partial E_2}{\partial w^{(i)}}\right)
\label{eq:update_w}
\end{equation}

\noindent where $\lambda$ is a scalar weight which helps to locally balance between accuracy and privacy. We update Alg. \ref{alg:client_bin} by adding an additional loss that helps further restrict leakage as indicated in Alg. \ref{alg:client_leak}.
\begin{algorithm}[t]
\SetAlgoLined
$s\leftarrow$ client socket initialized\\
$s.connectAndSyncParams(Server)$\\
\For{each batch $(x,y)$} {
    \textbf{Forward propagation:}\\
    $a_b^{(0)} \leftarrow x$\\
    \For{$i \leftarrow 1$ to $l$}{
        \textit{Same as in {\bf{Algorithm \ref{alg:client_bin}}}}
    }
    $a\red{'}_b^{(l)}\leftarrow a_b^{(l)}\red{+noise}$\\
    $s.send\left((a\red{'}_b^{(l)},y)\right)$\\
    \textbf{Backward propagation:}\\
    $\fracpartial{E}{a_b^{(l)}}\leftarrow s.receive()$\\
    \For{$i\leftarrow l$ downto $1$}{
        \textit{Same as in {\bf{Algorithm \ref{alg:client_bin}}}}
    }
}
$s.close()$
\caption{B-SL with DP integration (procedure on the client).}
\label{alg:client_dp}
\end{algorithm}

\subsection{Differential Privacy Integration}
In recent years, many researches study try to exploit DP to protect the privacy of SL such as \cite{SL1DCNN20Sharif,NotJustPrivacy18Wang,DPProtectSL21Ryu}. Therefore, DP could potentially be integrated into the proposed B-SL to mathematically guarantee privacy. Based on Eq. (\ref{eq:lap_noise}), in Alg. \ref{alg:client_dp} we integrate DP into B-SL (Alg. \ref{alg:client_bin}) by adding noise to the split layer's activations (${a'}_b^{(l)}\leftarrow a_b^{(l)}+noise$) before sending them to the server. However, due to binarization, smashed data is constrained to binary values instead of real values of noise added data. Therefore, instead of directly adding Laplace noise, we propose three different approaches to provide $\epsilon$-DP and also satisfy the binary output for smashed data as follows:

\begin{enumerate}
\item Stochastic binarization: instead of deterministic binarization, Shekhovtsov et al. \cite{StochasticBinarization21Shekhovtsov} proposed a stochastic binarization where they add a noise $z$ to real data such as $x_b=\textsf{Sign}(x-z)$. We could replace the noise $z$ by $\textsf{Lap}(\frac{s}{\epsilon})$ then the binarization $x_b'=\textsf{Sign}(x-\textsf{Lap}(s/\epsilon))$ provides $\epsilon$-DP where $s$ is the sensitivity of $x$. In order for this stochastic binarization to work, the authors proposed a new Straight Through Estimator which can be found in detail in \cite{StochasticBinarization21Shekhovtsov}. We formulate the equation for integrating DP into B-SL as follows.
\[{a'}_b^{(l)}\leftarrow\textsf{NewStraightThroughEstimator}(a_b^{(l)})\]

\item Double binarization: the idea of this second approach is simple and straightforward, that is to binarize a $\epsilon$-DP satisfied binarization to keep the output in form of binary values. The equation is as follows. Note that, based on post processing rule then ${a'}_b^{(l)}$ provides $\epsilon$-DP.
\[{a'}_b^{(l)}=\textsf{Sign}(a_b^{(l)}+\textsf{Lap}(2/\epsilon))\]

\item Randomizing response: this approach is motivated by a method of Warner et. al. \cite{RandomizedResponse65Warner} to improve bias in survey responses about sensitive issues. When a coin is flipped; if the coin is head, returns true value. If the coin is tail, flip another coin. If the second coin is head, return $1$; otherwise, return $-1$. With the probability $p$ of head or tail is $0.5$ then this mechanism satisfies $\epsilon$-DP with $\epsilon\approx1.04$ \cite{ProgrammingDP21Near}. In general, randomizing response satisfies $\epsilon$-DP with $\epsilon=\log(\frac{1+p}{1-p})$ where $p$ is the probability of the first coin is head. The equation for this approach is:
\[
    {a'}_b^{(l)}=
    \begin{cases}
        {a}_b^{(l)} \ \ \ \ \ \ \ & \text{with probability $p$}\\
            +1 & \text{with probability $(1-p)/2$}\\
            -1 & \text{with probability $(1-p)/2$}
    \end{cases}
\]
\end{enumerate}

In the next section, we will present the experimental results of these three approaches and suggest the better one in terms of less affecting model accuracy.
\section{Experimental evaluation} \label{sec:experiment}
We evaluate the performance of the proposed B-SL models in various classification tasks on 2D-image datasets. For privacy preservation evaluation, the correlation/similarity between raw and smashed data is used to measure leakage with various metrics such as mean square error, Kullback–Leibler divergence (KL-D) \cite{DroppingActivations18Dong}, distance correlation (DCOR) \cite{SL1DCNN20Sharif,NoPeek20Vepakomma}, structural similarity index measure (SSIM), and peak signal-to-noise ratio (PNSR) \cite{InversionAttacks19He}. In this work, KL-D and SSIM are chosen as the main metrics. While KL-D measures the difference between two probability distributions, SSIM measures the human perceptual similarity of two images and is used to demonstrate the visual reconstruction of the raw images from smashed data. Smaller distribution differences (smaller KL-D) and higher image similarities (larger SSIM) indicate high leaks and vice-versa.

\subsection{Experiment Setup}
We set up an SL network containing a client connected to a server as depicted in Fig. \ref{fig:SLArchitecture} where we localize the input and the first hidden layer while the rest are deployed in the server. The deep models are trained for $400$ epochs and an SGD optimizer \cite{SGDOverview16Ruder} with a batch size of $64$ is set. The SGD optimizer has the learning rate of $1e-2$, the momentum of $0.8$, and a weight decay of $5e-4$ by default. In addition, Dropout \cite{Dropout14Srivastava} is used as a regularization technique to reduce overfitting. We store the best-trained models based on validation loss which are then used to measure model accuracy on the test set.

\begin{table*}[!b]
\centering
\caption{Accuracy and privacy comparison between CNN, B-CNN and B-SL.}
\begin{tabular}{||c||c|c|c||c|c||c|c||}\hline
    Scheme & \textbf{B-CNN} & \textbf{B-SL} & \textbf{CNN} & \textbf{B-SL} & \textbf{CNN} & \textbf{B-SL} & \textbf{CNN} \\\hline
    \textit{Measure} & \multicolumn{3}{c||}{Accuracy ($\%$)} & \multicolumn{2}{c||}{KL-D ($\ge0$)} & \multicolumn{2}{c||}{SSIM ($[0,1]$)} \\\hline\hline
    \textit{MNIST} & $96.8$ & \textbu{$99.2$} & \textbu{$99.2$} & \textbu{$3.7$} & $1.5$ & \textbu{$0.2$} & \textbu{$0.2$} \\\hline
    \textit{Fashion MNIST} & $82.8$ & $90.2$ & \textbu{$90.9$} & \textbu{$3.3$} & $1.8$ & \textbu{$0.1$} & \textbu{$0.1$} \\\hline
    \textit{SVHN} & $72.9$ & $90.0$ & \textbu{$90.4$} & \textbu{$2.6$} & $1.1$ & \textbu{$0.1$} & \textbu{$0.1$} \\\hline
    \textit{CIFAR-10} & $70.5$ & $83.6$ & \textbu{$83.8$} & \textbu{$2.4$} & $0.9$ & \textbu{$0.1$} & \textbu{$0.1$} \\\hline
    \multicolumn{8}{l}{\textit{Note: higher KL-D and smaller SSIM values indicates lower leakage.}}
\end{tabular}
\label{tab:Accuracy_Leakage_Comparison}
\end{table*}

\textit{Datasets and deep models.} Four datasets with different complexities and domains have been used in the experiments: MNIST \cite{MNIST12Deng}, Fashion-MNIST \cite{FashionMNIST17Xiao}, SVHN \cite{SVHN11Netzer}, and CIFAR-10 \cite{CIFAR1009Krizhevsky}. These datasets contain grey-scale (MNIST and Fashion-MNIST) or color (SVHN and CIFAR-10) images in $10$ classes. To focus more on privacy preservation, data augmentation or pre-processing is not employed in these datasets. Regarding the deep models, our experiments evaluated LeNet-5 for MNIST and VGG-small \cite{ReviewDNN17Rawat} for the rest datasets. For easy referencing, we refer to all CNN based full-precision models as CNN; B-CNN is the complete binarization (binarize every convolution and fully connected layers, except the input and output layers) of the corresponding CNN; and B-SL is the proposed binarized SL model. 

\subsection{Performance Evaluation}
Table \ref{tab:Accuracy_Leakage_Comparison} presents the accuracy when conducting classification tasks on the four datasets. We conducted experiments and measured the accuracy of pure CNN-based models as a baseline for comparison with B-CNN and our B-SL models of the same architecture (same number of layers). We can see that B-CNN is less accurate than CNN due to data loss during binarization; the accuracy degradation is from $3\%$ to $27\%$ (e.g. $96.8\%$ compared to $99.2\%$ of CNN on MNIST or $70.5\%$ compared to $83.8\%$ of CNN on CIFAR-10).  Our proposed model, B-SL, achieves far better results than B-CNN and is comparable to CNN with less than $1\%$ difference in accuracy. For instance, the accuracy of B-SL is equal to CNN on MNIST while it is reduced by $0.7\%$, $0.4\%$ and $0.2\%$ respectively on Fashion MNIST, SVHN, and CIFAR-10 datasets.

In SL, the number of localized layers would have a significant impact on the local computation overhead of the client. Binarizing local layers would speed up the computation, but this could lead to more losses. In order to evaluate the effect of adding local layers to the model accuracy, we conducted experiments on CIFAR-10 dataset where the first, second, and third convolution layers of the VGG-small model are binarized and localized accordingly. The experiment results show that the accuracy is reduced from $83.6\%$ to $81.6\%$ and $81.1\%$ corresponding to splitting after one, two, and three convolution layers, respectively. By shifting a layer from the server to the client (further splitting point), we have more binarized layers (with more losses) and fewer full-precision layers that could explain the degraded accuracy of the entire model. However, in other experiments, we maintain the server-side model by distributing the first convolution layer to the client. Besides, at the client side we replicate this convolution layer multiple times. Then, the model accuracy is increased from $83.6\%$ to $84.7\%$ and $86.0\%$ corresponding to having one, two, and three local binarized convolutions, respectively. Therefore, adding more local binarized layers would increase or decrease the model accuracy depending on the splitting strategy (i.e. shifting or replicating layers). The addition of local layers definitely increases computation costs for the clients; however, regarding the effect of adding more layers to privacy leakage, we present the experimental results in the following subsection.

\begin{table*}[t]
\centering
\caption{Effects of adding additional leakage loss to the client compared to related work.}
\begin{tabular}{||c|c|c||c||c|c|c||}\hline
    \multirow{2}{*}{\textbf{Dataset}} & \multirow{2}{*}{\textbf{Model}} & \multirow{2}{*}{\textbf{\shortstack{Leakage restriction\\training approach}}} & \multirow{2}{*}{\textbf{\shortstack{Accuracy \\ ($\%$)}}} & \multicolumn{3}{c||}{Leakage measure} \\\cline{5-7}
    & & & & \textbf{KL-D} & \textbf{SSIM} & \textbf{DCOR} \\\hline\hline
    \textit{MNIST} & CNN & \cite{DroppingActivations18Dong} with $p=0.2$ & $99.1$ & $1.5$ & $0.2$ & $0.7$ \\\hline
    \textit{MNIST} & CNN & \cite{StepwiseActivations19Yu} with $n=21$ & $99.0$ & $1.7$ & $0.2$ & $0.8$ \\\hline
    \textit{MNIST} & CNN & Ours, DCOR loss \cite{NoPeek20Vepakomma} & $99.2$ & $1.7$ & $0.2$ & $0.8$ \\\hline
    \textit{MNIST} & CNN & Ours, SSIM loss & $99.4$ & $2.3$ & \textbu{$0.1$} & $0.8$ \\\hline
    \textit{MNIST} & B-SL & Ours, SSIM loss & $99.1$ & \textbu{$3.1$} & \textbu{$0.1$} & \textbu{$0.5$} \\\hline\hline
    \textit{CIFAR-10} & CNN & \cite{DroppingActivations18Dong} with $p=0.1$ & $83.2$ & $0.9$ & \textbu{$0.1$} & $0.7$ \\\hline
    \textit{CIFAR-10} & CNN & \cite{StepwiseActivations19Yu} with $n=231$ & $83.1$ & $0.9$ & \textbu{$0.1$} & $0.7$ \\\hline
    \textit{CIFAR-10} & CNN & Ours, DCOR loss \cite{NoPeek20Vepakomma} & $83.1$ & $0.7$ & \textbu{$0.1$} & \textbu{$0.6$} \\\hline
    \textit{CIFAR-10} & CNN & Ours, SSIM loss & $83.9$ & $0.8$ & \textbu{$0.1$} & \textbu{$0.6$} \\\hline
    \textit{CIFAR-10} & B-SL & Ours, SSIM loss & $83.8$ & \textbu{$2.9$} & \textbu{$0.1$} & $0.8$ \\\hline
\end{tabular}
\label{tab:Leakage_Comparison_SSIM}
\end{table*}

\subsection{Privacy Evaluation}

\begin{figure}[h]
    \centering
    \includegraphics[width=0.4\textwidth]{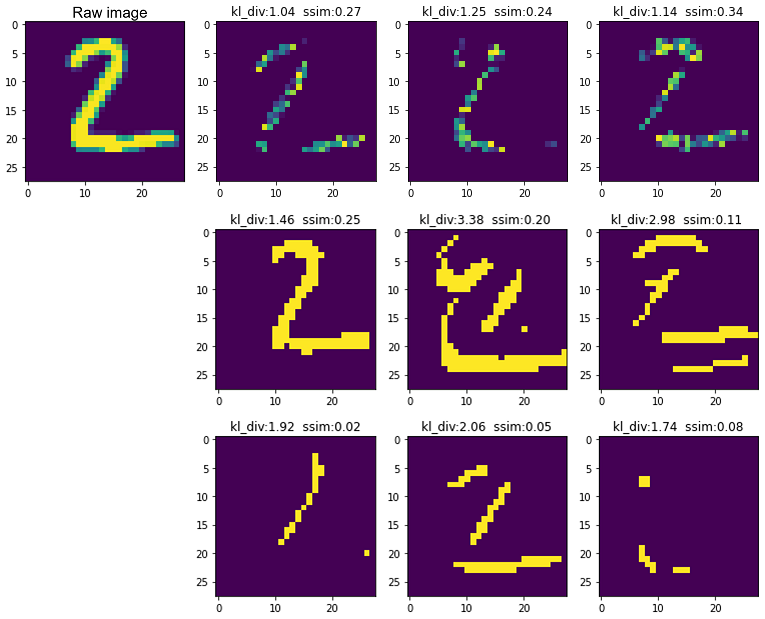}
    \caption{Visualizing activations from non-binarization (upper row), binarization (middle row) convolutions, and binarization with leakage restriction training (lower row). Raw input is on the top left.}
    \label{fig:Fashion_Leakage}
\end{figure}

\paragraph{With binarization} We measure raw data leakage to evaluate the privacy preservation of our B-SL compared to pure CNN. In order to isolate the effect of subsampling (e.g. max-pooling layers), we extract outputs from the last convolution layer towards the split layer for measuring leakages. We first visually reconstruct the 2D images from the activations and then compare them to the original image inputted to the model. Fig. \ref{fig:Fashion_Leakage} demonstrates images visually plotted from activations at the split layer of a CNN and B-SL with one localized convolution. We observe that binarization destroys color space and also maintains the partial shape of raw input which is similar to non-binarization. Similar to the accuracy experiments, we measured leakage of CNN and B-SL models over four datasets. The mean values of leakage from all channels are presented in Table \ref{tab:Accuracy_Leakage_Comparison}. Because of pixels pattern distorted by binarization, KL-D measurement of B-SL is outperform CNN. Note that, a higher KL-D value means greater dissimilarity and the KL-D values of B-SL are almost $2$ times higher than CNN. Besides, SSIM measurements reveal that CNN and B-CNN both maintain image structure and these models have equal results as 0.2 with MNIST and 0.1 on other datasets. SSIM scalar values are in the range $[0,1]$ where $1$ presents the most similar and $0$ is the least similar.

\paragraph{With leakage restriction training} In the experiments related to the proposed leakage restriction training approach, we implement SSIM as an additional leak loss and based on experimental results select the appropriate weight between effectiveness and privacy to balance the model's high accuracy while reducing leakage. Observing the experimental results presented in Table \ref{tab:Leakage_Comparison_SSIM}, we find that the model accuracy remains high (within $\pm0.5\%$ differs) while the SSIM values are further reduced up to $50\%$ in the case of evaluating both CNN and B-SL. For example, with MNIST experiments, the SSIM value is reduced from $0.2$ to $0.1$. The results from Table \ref{tab:Leakage_Comparison_SSIM} show that our proposed scheme achieves higher KL-D and smaller SSIM compared to other targets, including the work of \cite{DroppingActivations18Dong}, \cite{StepwiseActivations19Yu}, and \cite{NoPeek20Vepakomma}. We also visually reconstructed images from the split layer's activations of B-CNN after optimizing leakage and observe that there are some filters set to zero in order to minimize the leak loss. Zero-value filters could be removed from the model to make it lighter and therefore reduce computation overhead.

\begin{figure}[h]
    \centering
    \subfloat[Visualizing activations at split layer under DP protection with different epsilons.]{
        \includegraphics[width=0.4\textwidth]{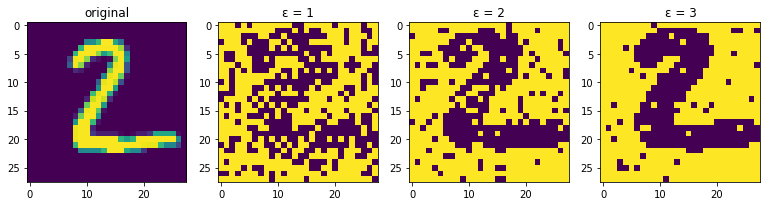}
    }
    \hfill
    \subfloat[Accuracy changes after applying DP with different $\epsilon$ among DP integration approaches for B-SL on MNIST dataset.]{
        \includegraphics[width=0.4\textwidth]{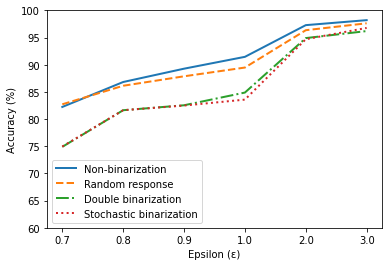}
    }    
    \caption{Effects of applying $\epsilon$-DP on split layer of B-SL.}
    \label{fig:DP_MNIST}
\end{figure}

\paragraph{With differential privacy} In order to evaluate the performance of the three proposed approaches for integrating DP into B-SL, we conducted experiments on the MNIST dataset where we run Alg. \ref{alg:client_dp} with stronger $\epsilon$ values. For each $\epsilon$ value, we sequentially apply stochastic binarization (SB), double binarization (DB), and randomizing response (RR) to the split layer's activations of B-SL then measure the affected accuracy of the models. We also measure the effect of $\epsilon$ on the accuracy of CNN (non-binarization) as the baseline for comparison. The results are plotted in Fig. \ref{fig:DP_MNIST} which reveals that the RR approach provides the smallest accuracy trade-off compared to SB and DB, and is comparable to the non-binarization model. Based on the experimental results, we could recommend that providing $\epsilon$-DP for B-SL by randomizing response would achieve better accuracy results. 

To evaluate the effect of adding more local layers to privacy preservation, we conducted experiments on both shifting layers from the server to the client and replicating the local layer. Experimental results are presented in Table \ref{tab:Adding_Layers_Comparison} where B$X$-SL is B-SL with $X$ first convolutions are deployed at the client and $X$B-SL is B-SL with the first local convolution replicates $(X-1)$ times. Under this naming convention, B1-SL and 1B-SL are examples of B-SL that are used in other experiments. The results reveal that adding more local layers slightly reduces leakage (such as increasing KL-D from $2.4$ to $2.9$). Moreover, shifting layers reduces more leakage than replicating layers because the latter convolutions usually produce smaller feature maps. However, the leakage reduction is not commensurate with the computation costs as adding one more layer would result in double overheads, especially with latter convolutions that have a high number of filters. In summary, considering accuracy, we suggest limiting the number of local layers to $2$ (by replicating the first convolution) to obtain higher results of both accuracy and privacy.

\begin{table}[t]
\centering
\caption{Leakage comparison when adding local layers in experiments with CIFAR-10 dataset.}
\begin{tabular}{||c||c|c|c||c|c|c||}\hline
    Metric & \multicolumn{3}{c||}{KL-D} & \multicolumn{3}{c||}{SSIM} \\\hline\hline
    \textbf{Scheme} & \textbf{B1-SL} & \textbf{B2-SL} & \textbf{B3-SL} & \textbf{B1-SL} & \textbf{B2-SL} & \textbf{B3-SL} \\\hline
    Measure & $2.35$ & $2.64$ & $2.58$ & $0.12$ & $0.08$ & $0.05$ \\\hline\hline
    \textbf{Scheme} & \textbf{1B-SL} & \textbf{2B-SL} & \textbf{3B-SL} & \textbf{1B-SL} & \textbf{2B-SL} & \textbf{3B-SL} \\\hline
    Measure & $2.35$ & $2.81$ & $2.87$ & $0.12$ & $0.11$ & $0.09$ \\\hline
\end{tabular}
\label{tab:Adding_Layers_Comparison}
\end{table}

Last, the binarization of weights, especially convolutional parameters, would lead to duplication of kernels. Duplicated kernels cause redundant computation that can be removed to reduce computational overhead \cite{DuplicateFilters17Roy}. Considering experiments on CIFAR-10 dataset, for example, the localized convolution has $64$ kernels of size $3\times3$. If we resize these kernels to $2\times2$, then there are possible $2^4$ different binary kernels which means high duplication during the model training. We performed again the experiments on CIFAR-10 with B-SL that has the localized convolution with $64$ of $2\times2$ kernels. The experiment results show that there are $42$ duplicated kernels over $64$ in total (corresponds to $66\%$). The same experiment on CNN models has shown that there is no duplication. Thus, binarization helps to reduce computation and cause duplicated kernels, which means a high possibility of reducing more computation costs that are a great advantage on mobile or IoT devices.

\subsection{B-SL Against Feature-Space Hijacking Attack} 
Feature-space hijacking attack (FSHA) \cite{FSHA_SL21Pasquini} is a novel attack that could lead to rethinking the SL protocol as the privacy-preserving properties are violated. Unlike the previous honest-but-curious server threat model, FSHA assumes a threat model with a dishonest server that actively hijacks the local private model by providing intentional gradients which combine true gradients (for model utility) and malicious gradients for mapping the local model to a pilot model. Then, the server could use the pilot model to reconstruct raw private data on receiving smashed data from the client. In \cite{FSHA_SL21Pasquini}, the authors demonstrated the effects of FSHA on SL, SL's variants such as SplitFed \cite{SplitFed21Thapa}, and defensive techniques such as NoPeek \cite{NoPeek20Vepakomma}. In our experiment, we study the defensive effectiveness of B-SL against FHSA.

\begin{figure}[h]
    \centering
    \includegraphics[width=0.4\textwidth]{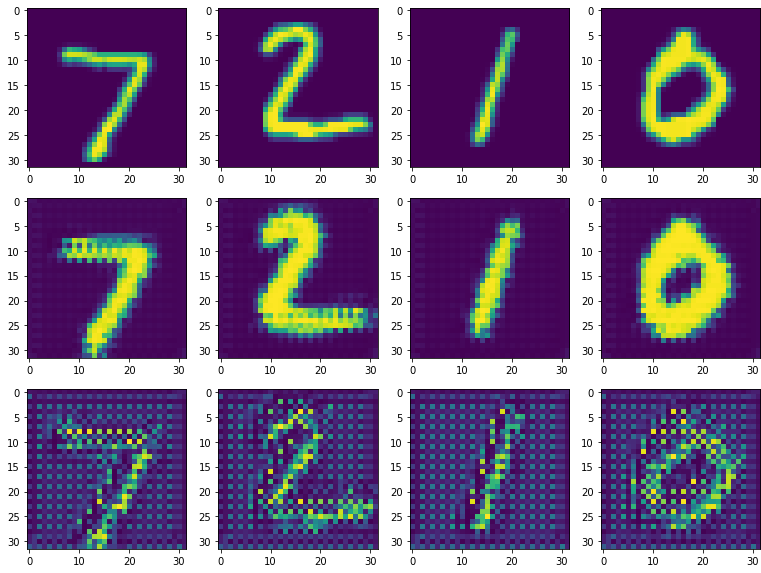}
    \caption{Images reconstruction when adopting FSHA on SL (middle row) and B-SL (bottom row) for the split 1. Top row shows original images.}
    \label{fig:FSHA_Split1}
\end{figure}

\begin{figure}[h]
    \centering
    \includegraphics[width=0.4\textwidth]{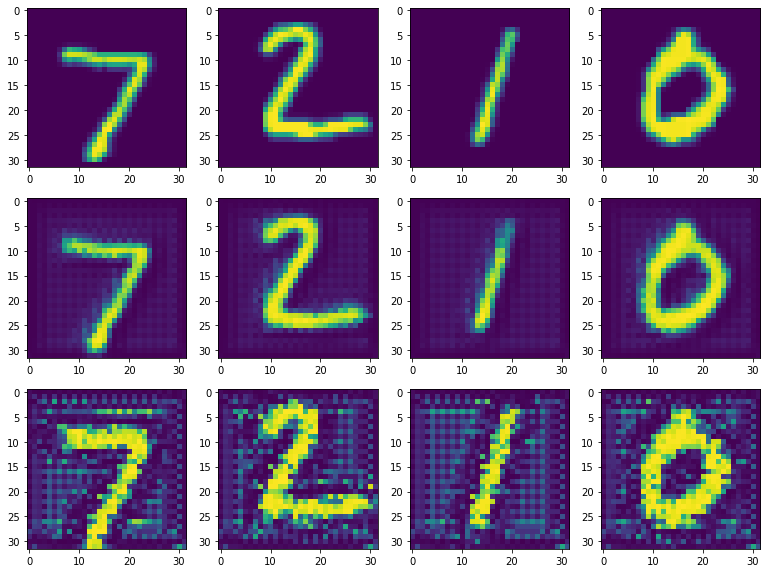}
    \caption{Images reconstruction when adopting FSHA on SL (middle row) and B-SL (bottom row) for the split 2. Top row shows original images.}
    \label{fig:FSHA_Split2}
\end{figure}

We setup an attack scenario that follows split $1$ and $2$ as presented in \cite{FSHA_SL21Pasquini} where we localize one and two convolutions, respectively. We apply the same architecture for attacker's models (i.e. $\tilde{f}$, $\tilde{f}^{-1}$, and $D$) as suggested in \cite{FSHA_SL21Pasquini}. Figs. \ref{fig:FSHA_Split1} and \ref{fig:FSHA_Split2} demonstrate the reconstruction results when FSHA is adopted to SL and B-SL. Visually, it can be observed that FSHA could successfully reconstruct raw data from SL's learning process. Regarding B-SL, interestingly, binarization helps relax this vulnerability. In order to numerically evaluate and compare B-SL to SL, we feed forward the reconstructed images from FSHA to the deep models to obtain the classification results. Besides the reconstruction loss (using MSE) between reconstructed and raw images, high classification accuracy results from reconstructed images indicates a high reconstruction quality. Without FSHA, both SL and B-SL models perform $99\%$ classification accuracy on MNIST dataset. Table \ref{tab:FSHA_Comparison} presents the experimental results when FSHA is adopted. For split 1, when adopting FSHA, the accuracy of SL is slightly reduced by $2\%$ (from $99\%$ without FSHA to $98\%$) while B-SL accuracy is more reduced by $3\%$ (from $99\%$ to $94\%$). The error of reconstruction which is measured by MSE loss between SL and B-SL are equal $0.7$; however, the quality of reconstruction images from SL is visually much better than from B-SL as shown in Fig. \ref{fig:FSHA_Split1}. Because of lower reconstruction quality, the accuracy of classification task on images reconstructed from B-SL is much lower than from SL, which are $10\%$ and $71\%$ respectively. When we increase the number of local layers to $2$ (split $2$), the quality of reconstruction is better than split 1 (see Fig. \ref{fig:FSHA_Split2}) and the accuracy of the classification on reconstructed images is also improved for both SL and B-SL (see Table \ref{tab:FSHA_Comparison}). Note that, B-SL still preserve better data privacy with reconstruction accuracy is $64\%$ compared to $96\%$ of SL.

\begin{table}[t]
\centering
\caption{Performance of FSHA on SL and B-SL over MNIST dataset.}
\begin{tabular}{||c|c|c|c|c||}\hline
    Scenario & Scheme & \shortstack{Model\\Accuracy} & \shortstack{Reconstruction\\Loss (MSE)} & \shortstack{Reconstruction\\Accuracy} \\\hline\hline
    \multirow{2}{*}{Split 1} & SL & $97\%$ & $0.7$ & $71\%$ \\\cline{2-5}
    & B-SL & $94\%$ & $0.7$ & $10\%$ \\\hline\hline
    \multirow{2}{*}{Split 2} & SL & $98\%$ & $0.7$ & $96\%$ \\\cline{2-5}
    & B-SL & $95\%$ & $0.6$ & $64\%$ \\\hline\hline
\end{tabular}
\label{tab:FSHA_Comparison}
\end{table}

On the experimental results related to FSHA, we could observe that B-SL helps relax the vulnerability of SL which is claimed to force rethinking the protocol \cite{FSHA_SL21Pasquini}. Regarding B-SL, when FSHA is adopted, the model accuracy drop is larger (than SL), an indicator for detecting the attack. Besides, when the local model is thin (less number of layers on the client-side), the effectiveness of FSHA is significantly reduced. These results could draw to insight for future investigations to study FSHA detection and defending.
\section{Conclusion} \label{sec:conclusion}
We have investigated the practicality of binarizing split learning to facilitate deep learning applications on mobile and IoT devices. We have introduced a hybrid deep model for collaboratively distributed learning where we only binarized the localized part while the server part is kept intact. We have proposed a leakage restriction training approach in order to further reduce leakage  from the shared smashed data while balancing the high classification accuracy of the models. To evaluate the proposal, we extensively conducted experiments on four commonly benchmarked datasets. The results demonstrate that adopting binarization helps lighten the localized model, speed up computation (up to $17.5$ forward-propagation times), reduce memory usage, and save data transmission bandwidth (up to $32$ times). Binarization models allow better privacy preservation in terms of KL-D and SSIM metrics compared to other targets experimentally. Moreover, training with leakage restriction further reduces leakage (measured by SSIM metric) up to $50\%$ and this method gets better results in binarization models. Finally, we could fully provide $\epsilon$-differential privacy for binarization models with the suggested randomizing response method to achieve the least accuracy trade-off. Taking these into account, our proposal would enable high-performance deep learning in an IoT environment with the maximum benefits at the inference stage. In future work, end-to-end evaluation in low-end devices should be thoroughly studied; and evaluating data leaks with other metrics on large-scale datasets should be further investigated.
%
%
\bibliographystyle{IEEEtran}
\bibliography{mybibliography}


\end{document}